\documentclass[a4paper,conference]{IEEEtran}
\usepackage{ifpdf}
\usepackage{cite}
\usepackage{enumitem}
\usepackage{tabularray}

\usepackage{graphicx}
\ifCLASSINFOpdf
\else
\fi
\usepackage{textpos}

\usepackage{amsfonts}
\usepackage{amsmath, bm}
\usepackage{algorithmic}
\usepackage{multirow}
\usepackage{array}
\ifCLASSOPTIONcompsoc
  \usepackage[caption=false,font=normalsize,labelfont=sf,textfont=sf]{subfig}
\else
  \usepackage[caption=false,font=footnotesize]{subfig}
\fi
\usepackage{stfloats}
\usepackage{url}

\newcommand{\mbf}[1]{{\mathbf{#1}}}
\newcommand{\refsec}[1]{Sec.~\ref{sec:#1}}
\newcommand{\reffig}[1]{Fig.~\ref{fig:#1}}
\newcommand{\refeq}[1]{Eq.~\ref{eq:#1}}
\newcommand{\reftab}[1]{Table~\ref{tab:#1}}

\newcommand{\ie}[1][ ]{{\em i.\thinspace{}e\@.{}}#1}
\newcommand{\eg}[1][ ]{{\em e.\thinspace{}g\@.{}}#1}
\newcommand{\ea}{{\em et al.}}

\DeclareMathOperator*{\argmax}{arg\,max}
\newcommand{\pose}[2]{^{#1}\mathbf{T}_{#2}}
\newcommand{\img}[1]{\mathbf{I}_{#1}}
\newcommand{\pt}{\mathbf{p}}

\begin{document}
\title{Forecasting of depth and ego-motion with transformers and self-supervision}
\author{
\IEEEauthorblockN{Houssem eddine BOULAHBAL}
\IEEEauthorblockA{Renault Software Factory and\\CNRS-I3S, \\Côte d'Azur University\\
Houssem-eddine.Boulahbal@renault.com}
\and
\IEEEauthorblockN{Adrian VOICILA}
\IEEEauthorblockA{Renault Software Factory\\
Adrian.Voicila@renault.com }
\and
\IEEEauthorblockN{Andrew I. COMPORT}
\IEEEauthorblockA{CNRS-I3S,\\Côte d'Azur University\\
Andrew.Comport@cnrs.fr}
}

\maketitle
\begin{abstract}
This paper addresses the problem of end-to-end self-supervised forecasting of depth and ego motion. Given a sequence of raw images, the aim is to forecast both the geometry and ego-motion using a self supervised photometric loss. The architecture is designed using both convolution and transformer modules. This leverages the benefits of both modules: Inductive bias of CNN, and the multi-head attention of transformers, thus enabling a rich spatio-temporal representation that enables accurate depth forecasting. Prior work attempts to solve this problem using multi-modal input/output with supervised ground-truth data which is not practical since a large annotated dataset is required. Alternatively to prior methods, this paper forecasts depth and ego motion using only self-supervised raw images as input. The approach performs significantly well on the KITTI dataset benchmark with several performance criteria being even comparable to prior non-forecasting self-supervised monocular depth inference methods.
\end{abstract}
\section{Introduction}

Forecasting the future is crucial for intelligent decision making. It is a remarkable ability of human beings to effortlessly forecast what will happen next, based on the current context and prior knowledge of the scene. Forecasting sequences in real-world settings, particularly from raw sensor measurements, is a complex problem due to the the exponential time-space space dimensionality, the probabilistic nature of the future and the complex dynamics of the scene. Whilst much effort from the research community has been devoted to video forecasting~\cite{Mathieu2016,Finn,WilliamLotter2016,Rakhimov} and semantic forecasting~\cite{Terwilliger2019,Bhattacharyya2019,Graber2021,Saric2020}, depth and ego-motion forecasting have not received the same interest despite their importance. The geometry of the scene is essential for applications such as planning the trajectory of an agent. 

Anticipating is therefore important for autonomous driving auto-pilots or human/robot interaction as it is critical for the agent to quickly respond to changes in the external environment. 
\let\thefootnote\relax\footnotetext{ This paper is a preprint (Accepted in ICPR 2022).\\
		\copyright 2022 IEEE. Personal use of this material is permitted. Permission from IEEE must be
		obtained for all other uses, in any current or future media, including
		reprinting/republishing this material for advertising or promotional purposes, creating new
		collective works, for resale or redistribution to servers or lists, or reuse of any copyrighted
		component of this work in other works.}

\begin{figure}
\centering
\includegraphics[width=0.44\textwidth]{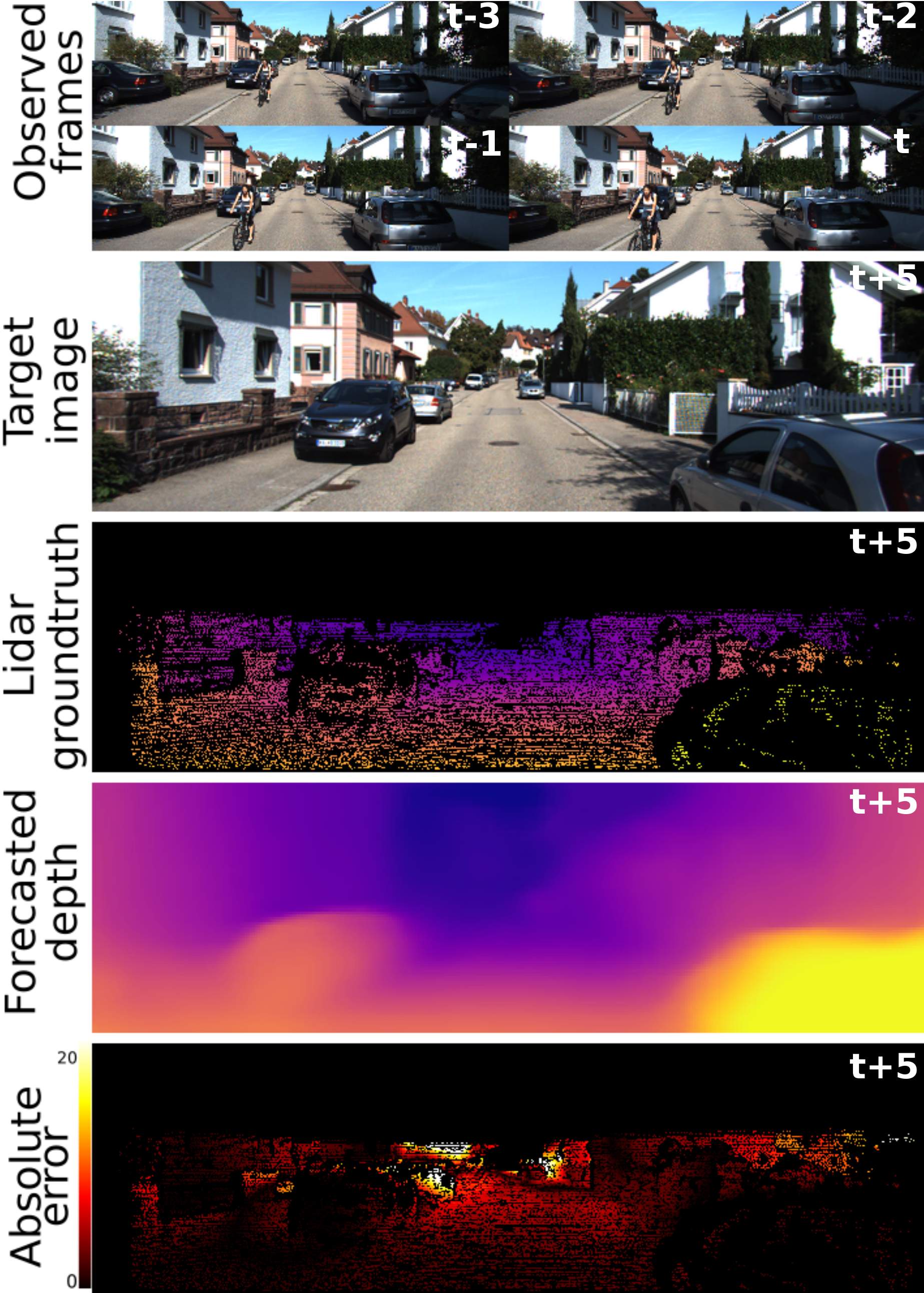}
\caption{The proposed method is trained using only raw sensor self-supervision, it is able to forecast an accurate geometry of the scene. The network only accesses the frames $\mbf{I}_{t-3:t}$ and forecast the depth $\mbf{D}_{t+n}$ and the pose $\pose{t}{t+n}$. Not only the network forecasts accurately the ego-motion (the depth of the static object is accurate) but also handles the dynamic objects.}
\label{fig:qualitative}
\end{figure}
The first work that studied depth forecasting was carried out by Mahjourian \ea~\cite{Mahjourian2017}, the aim of that paper was to use the forecasted depth to render the next RGB image frame. They supervised the depth loss using ground-truth LiDAR scans and the warping was done using ground-truth poses.~\cite{Qi2019} used additional modalities for input, namely, a multi-modal RGB, depth, semantic and optical flow and forecasted the same future modalities. The supervision was carried out using the aforementioned ground-truth labels.~\cite{Hu2020} developed a probabilistic approach for forecasting using only input images and generated a diverse and plausible multi-modal future including depth, semantics and optical flow. However, it was supervised through ground-truth labels and the final loss was a weighted sum of future segmentation, depth and optical flow losses similar to~\cite{Qi2019}. While these methods enable forecasting the depth, they suffer from two shortcomings:~\cite{Hu2020,Qi2019,Mahjourian2017} require the ground-truth labels for supervision during training and testing and~\cite{Qi2019} uses a multi-modal input for inference that requires either ground-truth labels or a separate network. 

The work presented in this paper addresses the problem of depth and ego-motion forecasting using only monocular images with self-supervision. Monocular depth and ego-motion inference has been successful for self-supervised training~\cite{Wang2021,Almalioglu2019,Jaderberg2015,Chen2016,Godard2017,Godard2019,Chen2019,Ranjan2019,Rares2020,Yin2018}. The basic idea is to jointly learn depth and ego-motion supervised by a photometric reconstruction loss. In this paper, it is demonstrated that it is possible to extend this self-supervised training to sequence forecasting. An accurate forecasting requires a knowledge of the ego-motion, semantics, and the motion of dynamic objects. Powered by the advances of transformers~\cite{Vaswani,Devlin2019,Brown2020,Dosovitskiy2020,Liu2021}, and using only sensor input, the network learns a rich spatio-temporal representation that encodes the semantics, the ego-motion and the dynamic objects. Therefore avoiding the need for extra labels for training and testing. The results on the KITTI benchmark~\cite{Geiger2012CVPR} show that the proposed method is able to forecast the depth accurately and outperform even non-forecasting methods~\cite{Eigen,Liu2016,Zhou2017,yang2018unsupervised}. 

\begin{figure*}
    \centering    
\includegraphics[width=\textwidth]{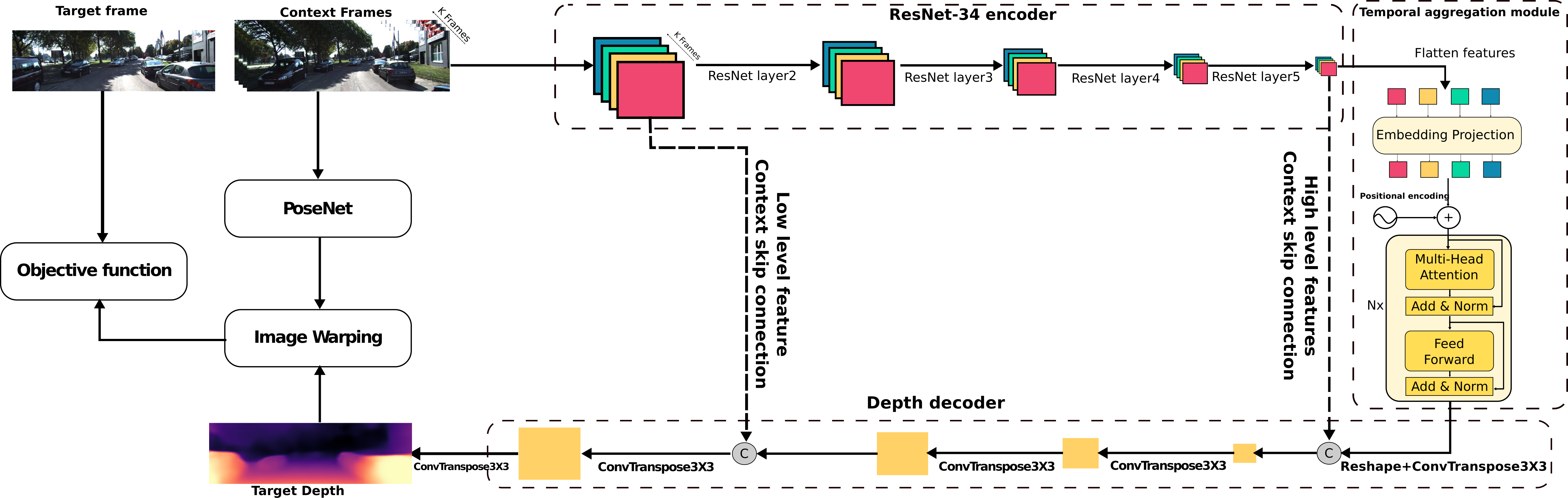}
    \caption{Illustration of the proposed architecture. Two sub-networks are used for training: The PoseNetwork as in network~\cite{Kendall2015,Godard2019} is used to forecast the ego-motion. The depth network combines both CNN and transformers. The Resnet34~\cite{he2016deep} encoder extracts the spatial features for each context frame. The embedding projection module projects these features into $R^{k \times d_{model}}$ where $k=4$ is the context frames. $N=3$ transformer encoders are used to fuse the spatial temporal to obtain a rich spatio-temporal features. The output of the transformer module encodes the motion of the scene. The decoder uses simple transposed convolution. In order to recover the context, skip connections are pooled from the encoder. Only the last frame features are pooled for the context. The decoder outputs a disparity map that will be used along with the pose network to warp the source images onto the target.}
    \label{fig:architecture}
\end{figure*}

\section{Related work}
Previous works~\cite{Godard2019,Zhou2017,Watson2021} have used the term "prediction" for methods that infer depth from a single image irrespective of past and future events. This could easily lead to confusion since the term "prediction" could equally be employed to refer to the prediction of future events. Therefore, the term "monocular depth inference" will be used to replace this ambiguous term. As the goal of this paper is related to time-series future prediction, the term "forecasting" will be used to refer to methods~\cite{Mahjourian2017,Qi2019,Hu2020} that use observations of the past and present to predict a future depth of the scene.

\subsection{Ego-motion and monocular depth self-supervised learning}
Monocular depth inference is an ill-posed problem as an infinite number of 3D scenes can be projected onto the same 2D scene. The earliest deep learning methods~\cite{Eigena,Eigen} utilized LiDAR data as ground truth for supervision. For inferring pose,~\cite{Kendall2015} proposed to use a simple regression network, based on a pretrained classifier~\cite{szegedy2015going}, supervised by an Euclidean loss. The work on spatial transformers~\cite{jaderberg2015spatial} offered an end-to-end differentiable warping function, thus, enabling several works~\cite{Zhou2017,Gordon2019,Zhan2018,Godard2019,Chen2019,Klingner2020,Bian2019,Watson2021,Wang2021,Rares2020,Ranjan2019} to formulate the problem as self-supervised training of depth and pose. To account for the ill-posed nature of this task, several works have addressed different challenges.~\cite{Bian2019,Chen2019,Rares2020,Wang2021} proposed to enforce depth scale and structure consistency.~\cite{Gordon2019,Shu2020} proposed more robust image reconstruction losses to handle occlusion due to moving objects.~\cite{Klingner2020,Yin2018,Ranjan2019,Chen2019} proposed multi-modal training to better supervise the network. In this paper, it will be shown that it is not only possible to infer the current depth but also forecast future depths accurately.

\subsection{Sequence forecasting}
Anticipation of the future state of a sequence is a fundamental part of the intelligent decision making process. The forecasted sequence could be a RGB video sequence~\cite{Finn,WilliamLotter2016,Babaeizadeh2018,Mathieu2016,Kumar2019,Rakhimov}, depth image sequence~\cite{weng2020inverting,Mahjourian2017}, semantic segmentation sequence~\cite{Terwilliger2019,Bhattacharyya2019,Luc2017,Graber2021,Chiu2020,Hu2020,Graber2021,Saric2020} or even a multi-modal sequence~\cite{Hu2020,Qi2019}. Early deep learning models for RGB video future forecasting leveraged several techniques including: Recurrent models~\cite{WilliamLotter2016}, variational autoencoder VAE~\cite{Babaeizadeh2018}, generative adversarial network~\cite{Mathieu2016}, autoregressive model~\cite{Rakhimov} and normalizing flows\cite{Kumar2019}. These techniques have inspired subsequent sequence forecasting methods.
Despite the importance of geometry for developing better decision making, depth forecasting is still in early development.~\cite{Mahjourian2017} used supervised forecasted depth along with supervised future pose to warp the current image and generate the future image. Instead of using images as input,~\cite{weng2020inverting} used LiDAR scans and forecast a sparse depth up to 3.0s in the future on the KITTI benchmark~\cite{Geiger2012CVPR}.~\cite{Qi2019} used a multi-modal input/output and forecast the depth among other modalities.~\cite{Hu2020} handled the diverse future generation by utilizing a variational model to forecast a multi-modality output. The use of multi-modalities requires additional labels or pretrained networks. This makes the training more complicated. 
Instead, the work presented in this paper leverages only raw images and forecasts in a self-supervised manner.

\subsection{Vision transformers}
The introduction of the Transformers in 2017~\cite{Vaswani} revolutionized natural language processing resulting in remarkable results~\cite{Devlin2019,Brown2020,radford2021learning}. The year 2020~\cite{Dosovitskiy2020,carion2020end} marked one of the earliest pure vision transformer networks. As opposed to recurrent networks that process sequence elements recursively and can only attend to short-term context, transformers can attend to complete sequences thereby learning long and short relationships. The multi-head attention could be considered as fully connected graph of the sequence's features. It demonstrated its success by outperforming convolution based networks on several benchmarks including classification~\cite{Dosovitskiy2020,zhai2021scaling,li2021improved}, detection~\cite{li2021improved,carion2020end,li2021grounded} and segmentation~\cite{Liu2021,cheng2021masked}. The has led to a paradigm shift~\cite{liu2021we}. transformers are slowly wining "The Hardware Lottery"~\cite{hooker2021hardware}. However, training vision transformers is complicated as these modules are not memory efficient for images and needs a large dataset pretraining.~\cite{carion2020end} has demonstrated that it is possible to combine convolution and transformers to learn a good representation without requiring large pertaining.
This paper proposes to leverage a hybrid CNN and transformer network as in~\cite{carion2020end} that is designed to forecast the geometry of the scene. The proposed network is simple and yet efficient. It outperforms even prior monocular depth inference methods~\cite{Eigen,Liu2016,Zhou2017,yang2018unsupervised} that access the image of the depth frame.

\section{The method}
\subsection{The problem formulation}
\label{sec:formulation}
Let $\mbf{I}_t \in \mathbb{R}^{w \times h \times c}$ be the t-th frame in a video sequence $\mbf{I} = \{ \mbf{I}_{t-k:t+n}\}$. The frames $\mbf{I}_c = \{\mbf{I_{t-k:t}}\}$ are the context of $\mbf{I}_{t}$ and $\mbf{I}_f =\{\mbf{I}_{t+1:t+n} \}$ is the future of $\mbf{I}_{t}$. 
The goal of the future depth and ego-motion forecasting is to predict the future geometry of the scene $\mbf{D}_{t+n}$ and the ego-motion $\pose{t+n}{t}$ corresponding to $\mbf{I}_{t+n}$ given only the context frames  $\mbf{I}_{c}$:
\begin{equation}\small
    \centering
    (\mbf{\widehat{D}}_{t+n}, ^{t+n}\mbf{\widehat{T}}_{t}) = f(\mbf{I}_c;\bm{\theta})
    \label{eq:problem_formulation}
\end{equation}
where $f$ is a neural network with parameters $\bm{\theta}$.

In self-supervised learning depth inference, the problem is formulated as novel view synthesis by warping the source frame $\mbf{I}_{s}$ into the target frames $\mbf{I}_{tar}$ using the depth and the $\pose{s}{tar} \in \mathbb{SE}[3]$ pose target to source pose. The warping is defined as: 
\begin{equation}\small\label{eq:warping}
        \widehat{\pt}_{s} = \pi ( \pose{s}{tar} H( \pi^{-1}(\pt_{tar}, D(\pt_{tar})))
\end{equation}
where $\pi$ is the inverse camera projection defined as:
\begin{equation}\small\label{eq:recover_z}
    \pi^{-1}\big(\pt, D(\pt)\big) = D(\pt)\Big(\frac{x-c_x}{f_x}, \frac{y-c_y}{f_y},1\Big)^\top
\end{equation}
$(f_x, f_y, c_x, c_y)$ are the camera intrinsic parameters. $\mbf{H}$ is the homogeneous coordinates transformation and $\pose{s}{tar} \in \mathbb{R}^{3 \times 4}$ is the ego-motion. Similarly, a pixel $\pt$ is projected to a 3D point $\mbf{P}$ given its depth $D(\pt)$ using the operator $\pi$ defined as:
\begin{equation}\small\label{eq:project}
    \pi(\pt) = (f_x \frac{X}{Z}+c_x, f_y \frac{Y}{Z} + c_y)
\end{equation}
using the spatial transformers~\cite{jaderberg2015spatial}, The reverse warping defined here is fully differentiable and uses a bilinear interpolation. Similarly, the self-supervised depth and ego-motion forecasting is defined here by considering the source frames $\mbf{I}_c$ and the target $\mbf{I}_{t+n}$ with the corresponding poses $\pose{t_i}{t+n} :\  \forall i \in \{t-k,...,t\}$. The frame $\widehat{\mbf{I}}_{t+n}$ is obtained by reverse warping the context frames.
  
Reconstructing the frame $\mbf{I}_{t+n}$ using the depth and the pose from only the context by a warping could be formulated as a maximum likelihood problem: 
 \begin{equation}
 \small
     \bm{\widehat{\theta}} = \argmax_{\bm\theta \in \bm{\Theta}} L(\mbf{I}_{t+n}| \mbf{I}_{c}; \bm{\theta} ) \equiv	 \argmax_{\bm\theta \in \bm{\Theta}} \sum_{m} P_{model}(\mbf{I}_{t+n}^m| \mbf{I}_{c}^m)
     \label{eq:likelihood}
 \end{equation} 
m is the number of samples. If $P_{model}$ is assumed to follow a Laplacian distribution $P_{model}(\mbf{I}_{t+n}| \mbf{I}_{c}) \sim Lap(\mbf{I}_{t+n};\bm{\mu}=\widehat{\mbf{I}}_{t+n};\bm{\beta}=\sigma^2 \mbf{I})$. $\widehat{\mbf{I}}_{t+n}$ is the warped image. Maximizing the~\refeq{likelihood} is equivalent to minimizing an $L_1$ error of $\mbf{\widehat{I}}_{t+n}$ and the known image frame $\mbf{I}_{t+n}$. Similarly, if the distribution is assumed to follow a Gaussian distribution, the maximization is equivalent to minimizing an $L_2$ error.

\subsection{The architecture}
The architecture of the network is depicted in \reffig{architecture}. The forecasting network is composed of two sub-networks: a pose net to forecast the future $\pose{t_i}{t+n}$ and a depth network that forecast $\mbf{D}_{t+n}$ (see~\refeq{problem_formulation})
Similar to \cite{Kendall2015}, the pose-net is composed of a classification network~\cite{he2016deep} as a feature extractor followed by a simple pose decoder as in~\cite{Godard2019}. The pose network forecasts 6 parameters using the axis-angle representation.
The depth network leverages a hybrid CNN and Transformer network as in~\cite{carion2020end} that is designed to forecast the geometry of the scene. This network benefits from both modules. The convolution module is used to extract the spatial features of the frames as it is memory efficient, easy to train and does not require large pretraining. The transformer module is used for better temporal feature aggregation. The multi-head attention could be considered as a fully-connected graph of the features of each frame. Therefore the information is correlated across all the frames rather than incrementally, one step at a time, as in LSTM~\cite{hochreiter1997long}. The architecture consists of three modules: an encoder, temporal aggregation module and a decoder.  
\subsubsection{Encoder: } 
ResNet~\cite{he2016deep} is one of the most used foundation models~\cite{bommasani2021opportunities}. It has demonstrated its success as a task agnostic feature extractor for nearly all vision tasks. In this work, ResNet34 is used as feature extractor. It is pretrained on ImageNet~\cite{deng2009imagenet} for better convergence. Each context frame is fed-forward and a pyramid of features is extracted. These features encode the spatial relationship between each scene separately. Thus, at the output of this module, a pyramid of spatial features for each frame is constructed. These features will be correlated temporally using the Temporal aggregation module TAM.

\subsubsection{Temporal aggregation module}
Since its introduction, transformer have demonstrated their performance outperforming their LSTM/RNN counterparts in various sequence learning benchmarks~\cite{Walker2021,raffel2019exploring,Brown2020,Devlin2019}. 
forecasting accurate depth requires knowledge of the static objects, accurate ego-motion and knowledge of the motion of the dynamic objects. The last layer of the encoder is assumed to encode higher abstraction features (\eg recognizing objects). Therefore, correlating temporally these features allows the extraction of the motion features of the scene. The TAM consists of two sub-modules: 
\begin{itemize}[leftmargin=*]
    \item \textbf{Embedding projection: } The dimensions after flattening the feature output of the last layer of the encoder is not memory efficient for the transformers. The embedding projection maps these features as:\footnote{The batch is omitted} $\mathbb{R}^{K \times C \times H \times W} \longrightarrow \mathbb{R}^{K \times d_{enc}}$. 
    \item \textbf{Transformer encoder: } After projecting the features using the embedding layer, a Transformer encoder with $N$ layer, $m$ multi-head attention and $d_{enc}$ is used. It correlates the spatial features of the sequence producing a spatio-temporal fused features.
\end{itemize}
\subsubsection{Depth decoder} After the spatio-temporal fusion, the decoder takes these spatio-temporal features along with the context features as input and decodes them to produce a disparity map. As depicted in~\reffig{architecture}, the context of the scene is obtained by pooling the features of last frame in the encoder. Two levels are pooled and concatenated. $^{dec}f_{t+n} = [^{enc}f_{t}  \: , \:  TAM(^{enc}f_{t-4:t})]$. The high level features (skip connection before the TAM) enable learning the motion while the low level features (skip connection at the start of ResNet) recover the finer details lost by the down-sampling. Therefore, the decoder maps (context + motion $\longrightarrow$ depth). 

Each level of the decoder consists of a simple sequential layer of: transposed convolution with kernel of $3\times3$ with similar channels as the encoder, batch normalization and Relu activation in that order. The forecasting head consists of a convolution with a kernel of $1\times1$ and a Sigmoid activation. The output of this activation, $\sigma$, is re-scaled to obtain the depth $D = \frac{1}{a \sigma +b}$, where $a$ and $b$ are chosen to constrain $D$ between $0.1$ and $100$ units, similar to~\cite{Godard2019}. For training, each level has a forecasting head but only the last head is used for inference. 

\subsection{Objective functions}
\label{sec:objetives}
As formulated in \refsec{formulation}, learning the parameters $\widehat{\bm{\theta}}$ involves maximizing the maximum likelihood of $P_{model}$. Similar to the prior work, along with the $L_1$ error other additional regularization terms are used and defined as:
\begin{itemize}[leftmargin=*]
    \item \textbf{Photometric loss: } Following~\cite{Zhou2017,Godard2019,Rares2020} The photometric loss seeks to reconstruct the target image by warping the source images using the forecast pose and depth. An $L_1$ loss is defined as follows:
    \begin{equation}\small
        \mathcal{L}_{rec}(\img{t+n}, \widehat{\mbf{I}}_{t+n}) = \sum_{\pt} | \img{t+n}(\pt) - \widehat{\mbf{I}}_{t+n}(\pt) |
    \end{equation}
    where $\widehat{\mbf{I}}_{t+n}(\pt)$ is the reverse warped target image obtained by \refeq{warping}. This simple $L_1$ is regularized using SSIM~\cite{wang2004image} that has a similar objective to reconstruct the image. The final photometric loss is defined as:
    \begin{equation}\small
    \begin{aligned}
      \mathcal{L}_\textup{pe}(\img{t+n}, \widehat{\mbf{I}}_{t+n}) = \sum_{\pt} \big[&(1 - \alpha) \textup{ SSIM}[ \img{t+n}(\pt) - \widehat{\mbf{I}}_{t+n}(\pt)]  \\
      & +  \alpha | \img{t+n}(\pt) - \widehat{\mbf{I}}_{t+n}(\pt) |\big]
    \end{aligned}
    \label{eq:ssim}
    \end{equation}
    \item \textbf{Depth smoothness: } An edge-aware gradient smoothness constraint is used to regularize the photometric loss. The disparity map is constrained to be locally smooth through the use an image-edge weighted $L_1$ penalty, as discontinuities often occur at image gradients. This regularization is defined as~\cite{heise2013pm}:
    \begin{equation}\small
    \begin{aligned}
      \mathcal{L}_{s}(D_{t+n}) = \sum_{p} \big[ &| \partial_x D_{t+n}(\pt) | e^{-|\partial_x \mbf{I}_{t+n}(\pt)|}    +\\
      &| \partial_y D_{t+n}(\pt) | e^{-|\partial_y \mbf{I}_{t+n}(\pt)|} \big] 
    \end{aligned}\label{eq:loss-disp-smoothness}
\end{equation}
\end{itemize}

Training with these loss functions is subject to major challenges: gradient locality, occlusion and out of view-objects. Gradient locality is a result of bilinear interpolation\cite{jaderberg2015spatial,Zhou2017}. The supervision is derived from the fours neighbors of $I(\pt_s)$ which could degrade training if that region is low-textured. Following~\cite{Godard2019,Godard2017,garg2016unsupervised}, an explicit multi-scale approach is used to allow the gradient to be derived from larger spatial regions. A forecasting head is used at each level to obtain each level's disparity map during training. 
\refeq{warping} assumes global ego-motion to calculate the disparity. Supervising directly using this objective is inaccurate when this assumption is violated (\eg the camera is static or a dynamic object moves with same velocity as the camera). According to~\cite{Godard2019} this problem can manifest itself as ‘holes’ of infinite depth. This could be mitigated by masking the pixels that do not change the appearance from one frame to the next.
A commonly used solution~\cite{Zhou2017,Godard2019} is to learn a mask $\mu$ that weigh the contribution of each pixel, while~\cite{Zhou2017} uses an additional branch to learn this mask. This paper uses the auto-masking defined in~\cite{Godard2019} to learn a binary mask $\mu$ as follows:
\begin{align}
\mu(\mbf{I}_{t+n}, \widehat{\mbf{I}}_{t+n}, \mbf{I}_{t}) =\mathcal{L}_{pe}(\img{t+n}, \widehat{\mbf{I}}_{t+n}) < \mathcal{L}_{pe}(\img{t+n},{\mbf{I}}_{t})
\end{align}
$\mu$ is set to only include the loss when the photometric loss of the warped image $\widehat{\mbf{I}}_{t+n}$ is lower than the original unwarped image $\img{t}$.
The final objective function is defined as:
\begin{equation}\small
    \mathcal{L} = \sum_{l} \:[ \:\mu \: \mathcal{L}_p + \alpha_d{L}_s \:]
\end{equation}
where $l$ is the scale level of the forecasted depth.

\section{Experiments}
\begin{table*}[t!]
\renewcommand{\arraystretch}{0.90}
\centering
{
\small
\setlength{\tabcolsep}{0.3em}
\begin{tabular}{c|ccc|cccc|ccc}
\hline
\textbf{Method} &
Forecasting &
Resolution & 
Supervision &
Abs Rel &
Sq Rel &
RMSE log &
RMSE &
$\delta<1.25$ &
$\delta<1.25^2$ &
$\delta<1.25^3$\\
\hline    
Eigen \ea~\cite{Eigen} &-& 576 x 271 & D  & 0.203 & 1.548 & 0.282 & 6.307 & 0.702 & 0.898 & 0.967 \\
Liu \ea~\cite{Liu2016}&-& 640 x 192 & D  & 0.201 & 1.584  & 0.273 & 6.471 & 0.680 & 0.898  & 0.967 \\
SfMLearner~\cite{Zhou2017} & -  & 416 x 128 & SS & 0.198 & 1.836& 0.275  & 6.565 & 0.718 & 0.901 & 0.960\\
Yang \ea~\cite{yang2018unsupervised} & - & 416 x 128 & SS & 0.182 & 1.481  & 0.267& 6.501 & 0.725 & 0.906 & 0.963\\
Vid2Depth~\cite{Mahjourian2018} & - & 416 x 128 & SS & 0.159 & 1.231  & 0.243 & 5.912& 0.784 & 0.923 & 0.970 \\
Monodepth2~\cite{Godard2019} & - & 640 x 192  & SS& 0.115 & 0.882 & 0.190 & 4.701 & 0.879 & 0.961 & 0.982\\
Wang \ea~\cite{Wang2021}  & - & 640 x 192  & SS & 0.109 & 0.779 & 0.186& 4.641 & 0.883 & 0.962 & 0.982\\
\hline
LiDAR Train set mean & -  & 1240 x 374& - & 0.361 & 4.826 & 0.377& 8.102  & 0.638 & 0.804 & 0.894\\
ForecastMonodepth2 & 0.5sec & 640 x 192 & SS & \underline{0.201}  &   \textbf{1.588}   &   \underline{0.275}  &   \textbf{6.166 }&   \underline{0.702}  &   \underline{0.897}  &   \underline{0.960}\\
\textbf{Ours}& 0.5sec & 640 x 192 & SS & \textbf{0.178}  &   \underline{1.645}   &   \textbf{0.257} &   \underline{6.196}  &   \textbf{0.761}  &   \textbf{0.914}  &   \textbf{0.964} \\
\hline
Copy last LiDAR scan & 1sec & 1240 x 374 & - &0.698  &  10.502  &  15.901  &   7.626  &   0.294  &   0.323  &   0.335\\
ForecastMonodepth2 & 1sec & 640 x 192&  SS& \underline{0.231}  &   \textbf{1.696}   &   \underline{0.303} &   \underline{6.685}  &   \underline{0.617}  &   \underline{0.869}  &   \underline{0.954} \\
\textbf{Ours}& 1sec &640 x 192 & SS & \textbf{0.208}  &   \underline{1.894}   &   \textbf{0.291 }&   \textbf{6.617}  &   \textbf{0.701}  &   \textbf{0.882}  &   \textbf{0.949}\\
\hline
\end{tabular}
}

\caption{Quantitative performance comparison of  on the KITTI benchmark with Eigen split~\cite{Geiger2012CVPR} for distances up to 80m.
In the \emph{Supervision} column,  D refers to depth supervision using LiDAR groundtruth and (SS) self-supervision. At test-time, all monocular methods (M) scale the depths with median ground-truth LiDAR.}
\label{tab:depth-accuracy}
\end{table*}

\begin{table}
    \centering 
    \begin{tabular}{c|c|c|c}
         Method & forecasting & Seq.09 & Seq.10 \\
         \hline
         Mean Odom & - &0.032 $\pm$ 0.026 & 0.028 $\pm$ 0.023\\
         ORB-SLAM~\cite{mur2015orb}  & - & 0.014 $\pm$ 0.008 & 0.012 $\pm$ 0.011\\ 
         SfMLearner~\cite{Zhou2017} & - & 0.021 $\pm$ 0.017 & 0.020 $\pm$ 0.015 \\
         Monodepth2~\cite{Godard2019} & - & 0.017 $\pm$ 0.008 & 0.015 $\pm$ 0.010 \\
         Wang \ea~\cite{Wang2021} &  - & 0.014 $\pm$ 0.008 & 0.014 $\pm$ 0.010 \\
         \hline
         Ours & 0.5s & 0.020 $\pm$ 0.011 & 0.018 $\pm$ 0.011 \\
         \hline
    \end{tabular}
    \caption{ATE error of the proposed method and the prior non-forecasting methods on KITTI~\cite{Geiger2012CVPR}. The proposed method is comparable to these methods even if it only accesses past frames.}
    \label{tab:pose}
\end{table}

\subsection{Setting}
\subsubsection{KITTI benchmark~\cite{Geiger2012CVPR}: } 
Following the prior work~\cite{Eigen,Liu2016,Zhou2017,yang2018unsupervised,Mahjourian2018,Godard2019,Wang2021}, the Eigen~\ea~\cite{Eigen} split is used with Zhou~\ea~\cite{Zhou2017} pre-processing to remove static frames. Frames without sufficient context are excluded from the training and testing. This split has become the defacto benchmark for training and evaluating depth that is used by nearly all depth methods.
\subsubsection{Baselines: }
\label{sec:baselines}
As discussed above, previous work on depth forecasting has been supervised using LiDAR scans, and has used a multimodal network that provides depth. Their evaluation is not performed on the Eigen split, nor does it use the defacto self-supervised metrics.
In order to fairly evaluate the proposed method, a self-supervised monocular formulation will be used to compare performance with the KITTI Eigen split benchmark. Comparisons will be made with three approaches: prior work on self-supervised depth inference~\cite{Eigen,Liu2016,Zhou2017,yang2018unsupervised,Godard2019,Wang2021}; copy of the last observed LiDAR frame as done in~\cite{Qi2019}; and ForecastMonodepth2, a modified version of~\cite{Godard2019} that is adapted for forecasting pose/depth.
\subsubsection{Evaluation metrics: }
For evaluation, the metrics of previous works~\cite{Eigen} are used for the depth. During the evaluation, the depth is capped to 80m. To resolve the scale ambiguity, the forecasted depth map is multiplied by median scaling where $s = \frac{median(D_{gt})}{median(D_{pred})}$. For the pose evaluation, the Absolute Trajectory Error (ATE) defined in~\cite{sturm2012benchmark} is evaluated on KITTI odometery benchmark~\cite{Geiger2012CVPR} sequences 09 and 10. 
\subsubsection{Implementation details: }
PyTorch~\cite{pytorch} is used for all models. The networks are trained for 20 epochs, with a batch size of 8. The Adam optimizer~\cite{kingma2014adam} is used with a learning rate of $lr=10^{-4}$ and $(\beta_1, \beta_2) =  (0.9,0.999)$. As training proceeds, the learning rate is decayed at epoch 15 to $10^{-5}$. The SSIM weight is set to  $\alpha = 0.15$ and the smoothing regularization weight to $\alpha_d = 0.001$. $l=4$ scales are used for each output of the decoder. At each scale, the depth is upscaled to the target image size. $d_{model}=2048$, $m=16$ and $N=3$ for the TAM projection.
The input images are resized to $192 \times 640$. Two data augmentations were performed: horizontal flips with probability $p=0.5$ and color jitter with $p=1$.
$k=4$ frames are used for context sequence and $n=5$ is used for short term forecasting and $n=10$ for mid-term forecasting as in~\cite{Qi2019} which corresponds to forecasting $0.5$s and $1.0$s into the future. 
The ForecastedMonodepth2 is the same as~\cite{Godard2019} with a modified input. The context images are concatenated and used as input for both depth and pose networks.

\subsection{Depth forecasting results}
\reftab{depth-accuracy} shows the results of the proposed method on the KITTI benchmark~\cite{Geiger2012CVPR}. As specified in \refsec{baselines}, the method is compared to three approaches: prior work on depth inference; copying last frame; and adapting monodepth2~\cite{Godard2019} for future forecasting. The proposed method outperforms the forecasting baselines for both short and mid-term forecasting especially for short range forecating. The results are even comparable to non-forecasting methods~\cite{Eigen,Liu2016,Zhou2017,yang2018unsupervised} that have access to $\mbf{I}_{t+n}$. The gap between state-of-the-art depth inference and the proposed forecasting method is reasonable due to the uncertainty of the future, the unobservability of certain events such as a new object entering the scene and the complexity of natural videos that requires modeling correlations across space-time with much higher input dimensions.

\reffig{qualitative} shows an example of depth forecasting on Eigen test split. Several observation can be made: 
\begin{itemize}[leftmargin=*]
    \item The network handles correctly the out-of-view object (\ie the bicycle on the scene).
    \item The network learned the correct ego-motion: The position of the static objects is accurate.
\end{itemize}
These results suggest that the network is able to learn a rich spatio-temporal representation that enable learning the motion, geometry and the semantics of the scene. Thus, extending the self-supervision depth inference to perform future forecasting with comparable results. A further analysis is done to evaluate and validate the choices of the network in~\refsec{ablation}.

\subsection{Ego-Motion forecasting results}
\reftab{pose} shows the results of the proposed network on the KITTI odometry benchmark~\cite{Geiger2012CVPR}. Similar to depth, the assessment is made by comparing with non-forecasting prior works. To avoid data leakage, the network is trained from scratch on the sequences 00-08 of the KITTI odometry benchmark. 
The network takes only the context images $\mbf{I}_c$ and forecasts $\pose{t}{t+n}$. The ATE results in \reftab{pose} show the proposed network achieves a competitive result relative to other non-forecasting approaches. All the methods are trained in the monocular setting, and therefore scaled at test time using ground-truth. These results suggest that using the proposed architecture along with the self-supervised loss function  successfully learns the future joint depth and ego-motion.

\begin{table}
\centering
{
\small
\setlength{\tabcolsep}{0.3em}
\begin{tabular}{c|cccc}
\hline
\textbf{Method} &
Abs Rel &
Sq Rel &
RMSE log & 
RMSE 
\\
\hline
\textbf{Ours} & 0.178  &   1.645    &   0.257&   6.196 \\
Without TAM & 0.205  &   1.745    &   0.296 &   6.565 \\
Shared pose/depth features & 0.208  &   1.745    &   0.282  &   6.529 \\
Single scale & 0.208  &   1.950  &   0.283   &   6.595  \\
Disable auto-masking &  0.193  &   1.774   &   0.273  &   6.374 \\
\hline
\end{tabular}
}
\caption{Ablation study of the different component of the proposed method. The evaluation was done on KITTI benchmark using Eigen split~\cite{Eigen}}
\label{tab:ablation}
\end{table}

\subsection{Ablation study}
\label{sec:ablation}
To further analyse the network, several ablations are made.~\reftab{ablation} depicts a comparison of the proposed model with several variants. The evaluation is done for short-term forecasting $n=5$ using $k=4$ context frames.

\subsubsection{Effect of the Temporal Aggregation Module} 
In order to evaluate the contribution of the multi-head attention, a variant of the proposed method is designed by replacing the TAM module by a simple concatenation of the last layer features. From \reftab{ablation}, the improvement induced by the TAM module is significant across all metrics. These results suggest that the performance obtained by the proposed method is achieved through the TAM module. Since the TAM aggregates the temporal information across all frames using a learned attention, the temporal features are better correlated and the final representation successfully encodes the spatio-temporal relationship between the images.

\subsubsection{Effect of sharing the encoder of depth and ego-motion}
Since both pose and depth networks encode the future motion and geometry of the scene, it is expected that sharing the encoders of these networks yield better results. However, as reported by~\reftab{ablation}, the degradation is significant. Even though these tasks are collaborative, sharing the encoder will result in a set of parameters $\hat{\bm{\theta}}$ that are neither the best local optima for the depth nor for the pose. By alleviating this restriction and separating the encoders, the network learns better local optima for both pose and depth.

\subsubsection{The benefit of using multiple scales} 
In order to evaluate the multi-scale extension, a variant of the proposed method that uses only one scale is trained. As illustrated in the \reftab{ablation} The network benefits from the multi-scale. The reverse warping 
uses bi-linear interpolation. As mentioned earlier, each depth point depends only on the four neighboring warped points. By using a multi-scale depth at training-time the gradient is derived from a larger spatial region directly at each scale. 

\subsubsection{Effect of auto-masking}
\reftab{ablation} compares the proposed method with a variant without using the auto-masking defined is~\refsec{objetives}. The results show that using auto-masking improves all four evaluation criteria. This demonstrates that using auto-masking, for pixels that do not change appearance, reject these outliers that inhibit the optimization. This leads to better accuracy of the forecasted depth.

\subsection{Limitations and perspective}
Even though the proposed forcasting method yields good results, there exists a gap with respect to non-forecasting methods. Several limitations contribute to this:
\begin{itemize}[leftmargin=*]
    \item A common assumption across SOTA methods is that the environment is deterministic and that there is only one possible future. However, this is not accurate since there are multiple plausible futures. Given the stochastic nature of the forecasting proposed here, the network will tend to forecast the mean of all the possible outcomes~\cite{Babaeizadeh2018}.  
    
    \item  The network does not forecast the correct boundaries of the objects. This is due to the formulation as a maximum likelihood problem with a Laplacien distribution assumption and the deterministic nature of the architecture. As a result, the boundaries of the dynamic objects are smoothed. 
    \item Due to the problem formulation, the scale of the forecasted depth is ambiguous. 
    \item The model fails to account for the motion of distant dynamic objects due to lack of parallax. and fine objects are ignored by the network.
\end{itemize}
To address these limitations future work will investigate constraining depth to be conditional on the input image structure as in~\cite{Boulahbal2021}. Concerning scale, one solution could be to infer directly from the scene as in~\cite{Bian2019,Chen2019,Wang2021}.

\section{Conclusion}
This paper proposed an approach for forecasting future depth and ego motion using only raw images as input. This problem is addressed as end-to-end self-supervised forecasting of the future depth and ego motion. Results showed significant performances on several KITTI dataset benchmarks~\cite{Geiger2012CVPR}. The performance criteria are even comparable with non-forecasting self-supervised monocular depth inference methods~\cite{Eigen,Liu2016,Zhou2017,yang2018unsupervised}. The proposed architecture demonstrates the effectiveness of combining the inductive bias of the CNN as a spatial feature extractor and the multi-head attention of transformers for temporal aggregation. The proposed method learns a spatio-temporal representation that captures the context and the motion of the scene. 
\footnote{This work was performed using HPC resources from GENCI-IDRIS (Grant 2021-011011931). The authors would like to acknowledge the Association Nationale Recherche Technologie (ANRT) for CIFRE funding (n°2019/1649).
}

\bibliographystyle{plain}
\bibliography{bib}
\newpage
\section{More Qualitative results}

\begin{figure*}
    \centering
    \includegraphics[width=\textwidth]{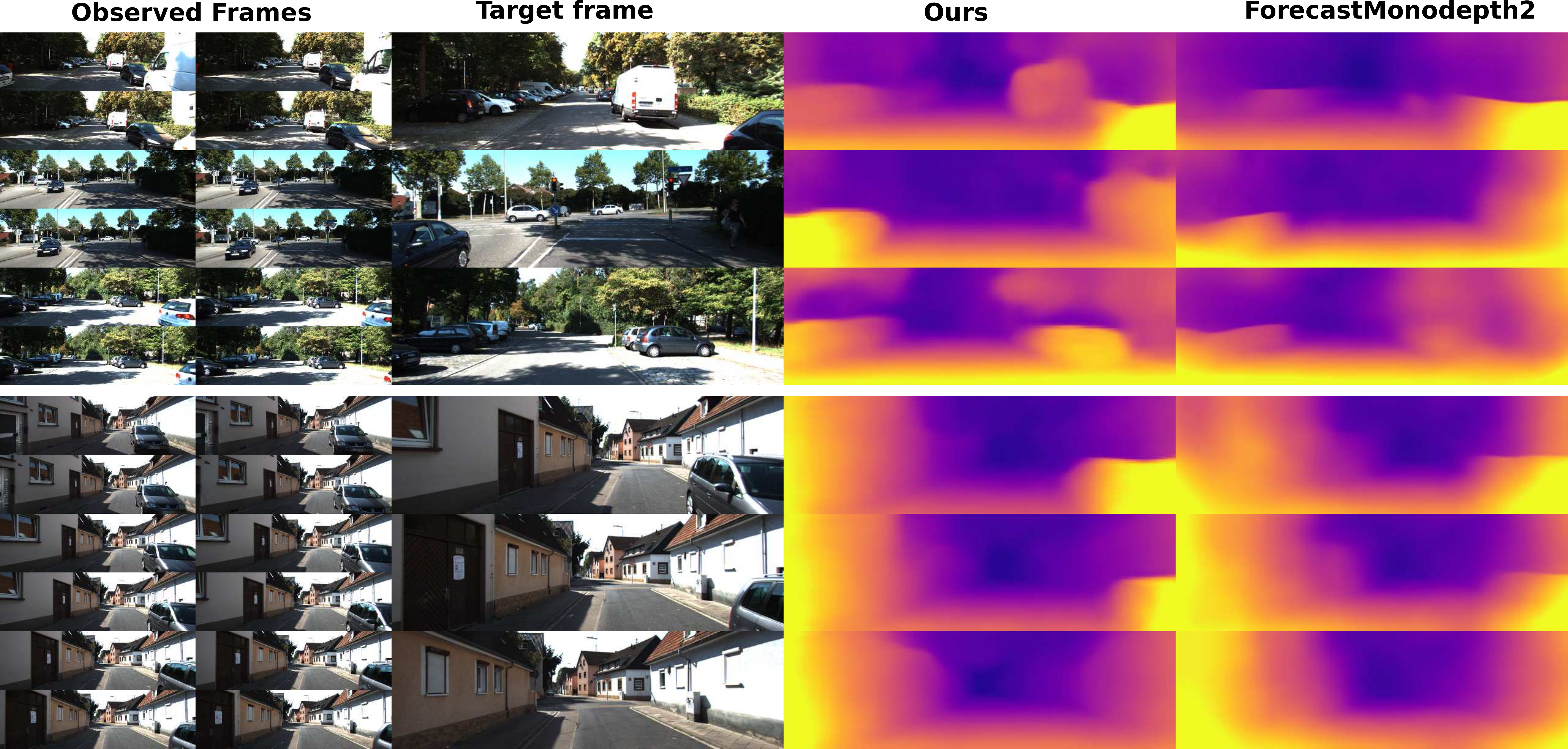}
    \caption{Qualitative results of the comparison of the proposed method with the ForecastMonodepth2 baseline. This comparison shows that the proposed method performs better than the baseline, especially for nearby dynamic objects. This observation is further validated in Table IV. In addition, the baseline method is showing a lack of detection of moving objects, which leads to a degradation of the forecasted depth. The proposed method is able to detect moving objects, thus accurately forecasting the depth of the scene.}
    \label{fig:qualitative_2}
\end{figure*}

\begin{table*}
\centering
\begin{tabular}{|c|cc|ccccccc|}
    \hline
    \textbf{Range} &
    Method &
    Forecasting &
    Abs Rel &
    Sq Rel &
    RMSE log &
    RMSE &
    $\delta<1.25$ &
    $\delta<1.25^2$ &
    $\delta<1.25^3$\\
    \hline
    \multirow{3}{*}{\textbf{[00m 10m]}}
    & Monodepth2~\cite{Godard2019} & - &  0.066  &   0.264  &  0.106& 1.076    &   0.959  &   0.987  &   0.994  \\
    & ForecastMonodepth2 & 0.5s &  0.138  &   \textbf{0.586} &   0.178 &   1.697 &   0.847  &   0.957  &   0.985  \\
    & \textbf{Ours} & 0.5s &   \textbf{0.112}  &   0.595  &  \textbf{0.155} & \textbf{1.573}    &   \textbf{0.893}  &   \textbf{0.964}  &   \textbf{0.986}  \\
    \hline
    \multirow{3}{*}{\textbf{[10m 30m]}}
    &Monodepth2~\cite{Godard2019} & - &  0.119  &   0.858   &   0.192  &   3.706 &   0.876  &   0.956  &   0.978   \\
    & ForecastMonodepth2 &0.5s&0.192  &   1.673  &   0.258  &   5.169   &   0.725  &   0.906  &   0.963  \\
    & \textbf{Ours} &0.5s&\textbf{ 0.167}  &  \textbf{ 1.453} &   \textbf{0.241 }&   \textbf{4.803}    & \textbf{  0.782}  &   \textbf{0.921}  &  \textbf{ 0.965}  \\
    \hline
    \multirow{3}{*}{\textbf{[30m 80m]}}
    &Monodepth2~\cite{Godard2019} & - &  0.188  &   3.094  &  11.115  &   0.288  &   0.709  &   0.897  &   0.950   \\
    & ForecastMonodepth2 &0.5s& \textbf{0.213}  &   \textbf{3.526}  &  \textbf{11.940}  &   \textbf{0.292 } &   \textbf{0.631}  &   \textbf{0.874}  &   \textbf{0.953}  \\
    & \textbf{Ours} &0.5s& 0.224  &   4.052  &  12.638  &   0.312  &   0.622  &   0.862  &   0.941  \\
    \hline
\end{tabular}
\caption{Quantitative performance comparison on the KITTI benchmark with Eigen split~\cite{Geiger2012CVPR} for multiple distances range. For Abs Rel, Sq Rel, RMSE and RMSE log lower is better, and for $\delta < 1.25$, $\delta < 1.25^2$ and $\delta < 1.25^3$ higher is better. Three ranges are considered: short range [0 10m] which represents 37.95\%, medium-range [10 30]which represents 50.74\% and long-range[30 80] which represents 11.30\%. The results shows that the proposed method is able to forecast good depth and outperform the baseline at short and medium forecasting range.}
\label{tab:distance_filtred}

\end{table*}

In order to further analyse the depth forecasting results, an assessment based on the ground-truth LiDAR distance is done.~Table IV shows the comparison of the non-forecasting method Monodepth2~\cite{Godard2019}, ForecastMonodepth2 and the proposed methods. 

The results suggest that the proposed method outperform the adaptation of Monodepth2 for short-range with a improvement of the Abs Rel of $-16.7\%$ and medium-range with a improvement of the Abs Rel of $-8.8\%$. These regions are the most significant regions of the forecasting as it has enough parallax for the ego-motion and dynamic object motion. Besides, this region assess several challenges including out-of-view objects and occlusion. For the long-range forecasting, the results shows that the two methods performer badly due to the lack of parallax in these region and down-sampling that ignore small objects. Moreover, this region has a high likelihood of new-object entering the scene which the forecasting is unable to handle by definition. The reported performances and the qualitative results suggest that the two forecasting networks only fit the road and completely ignore any other object. These results are shown qualitatively in \reffig{qualitative_2}.

\end{document}